\documentclass[conference]{IEEEtran}
\IEEEoverridecommandlockouts
\usepackage{cite}
\usepackage{amsmath,amssymb,amsfonts}
\usepackage{algorithmic}
\usepackage{graphicx}
\usepackage{textcomp}
\usepackage{xcolor}
\usepackage{xspace}
\usepackage{subfig}
\usepackage{tabu}
\usepackage{tabularx}
\usepackage{multirow}
\usepackage{breakurl}
\usepackage[breaklinks]{hyperref}
\hypersetup{hidelinks,
colorlinks=false,
}
\def\BibTeX{{\rm B\kern-.05em{\sc i\kern-.025em b}\kern-.08em
    T\kern-.1667em\lower.7ex\hbox{E}\kern-.125emX}}

\newcolumntype{L}[1]{>{\raggedright\let\newline\\\arraybackslash\hspace{0pt}}m{#1}}
\newcolumntype{C}[1]{>{\centering\let\newline\\\arraybackslash\hspace{0pt}}m{#1}}
\newcolumntype{R}[1]{>{\raggedleft\let\newline\\\arraybackslash\hspace{0pt}}m{#1}} 
\newcommand{\mdgan}{MD-GAN\xspace}
\newcommand{\mdsimple}{MD-GAN$^s$\xspace}
\newcommand{\mdswap}{MD-GAN$^w$\xspace}

\newcommand{\alg}{\textsc{DFG}\xspace}  
\newcommand{\defense}{\textsc{DFG}\xspace} 
\newcommand{\defenseadjust}{\textsc{DFG\_A}\xspace} 
\newcommand{\defenseadjustp}{\textsc{DFG\_A+}\xspace} 
\newcommand{\defensep}{\textsc{DFG+}\xspace} 
\newcommand{\defenseplus}{\textsc{DFG+}\xspace} 
\newcommand{\dft}{\textsc{DFG(+)}\xspace}
\newcommand{\mdsimplenodefense}{No\_Def\_Simple\xspace} 
\newcommand{\mdswapnodefensep}{No\_Def\_Swap\xspace}

\newcommand{\detector}{detector\xspace}

\begin{document}
\xspaceremoveexception{-}
\title{Attacks and Defenses for Free-Riders in Multi-Discriminator GAN}

\author{\IEEEauthorblockN{Zilong Zhao\IEEEauthorrefmark{1}\textsuperscript{\textsection}, Jiyue Huang\IEEEauthorrefmark{1}\textsuperscript{\textsection}, Stefanie Roos\IEEEauthorrefmark{1} , Lydia Y. Chen\IEEEauthorrefmark{1}}
\IEEEauthorblockA{\IEEEauthorrefmark{1}TU Delft, The Netherlands. \{Z.Zhao-8, J.Huang-4, S.Roos, Y.Chen-10\}@tudelft.nl} 
}

\maketitle
\begingroup\renewcommand\thefootnote{\textsection}
\footnotetext{Equal contribution}
\endgroup
\begin{abstract}
Generative Adversarial Networks (GANs) 
are increasingly adopted by the industry to synthesize realistic images using competing generator and discriminator neural networks. 
Due to data not being centrally available, Multi-Discriminator (MD)-GANs training framework employ multiple discriminators that have direct access to the real data. 
Distributedly training a joint GAN model entails the risk of free-riders, i.e., participants that aim to benefit from the common model while only pretending to participate in the training process.  
In this paper, we conduct the first characterization study of the impact of free-riders on \mdgan.
Based on two  production prototypes of \mdgan, 
we find that 
free-riders drastically reduce the ability of MD-GANs to produce images that are indistinguishable from real data, i.e., they increase the FID score --- the standard measure to assess quality of generated images.
To mitigate the model degradation, we propose a defense strategy against free-riders in \mdgan, termed \alg. \alg distinguishes free-riders and benign participants through periodic probing and clustering of discriminators' responses based on a reference response of a free-riders, which then allows the generator to exclude the detected free-riders from the training. Furthermore, we extend our defense, termed \defensep, to enable discriminators to filter out free-riders at the variant of \mdgan  that allows peer exchanges of discriminators networks.
Extensive evaluation on various scenarios of free-riders, \mdgan architecture, and three datasets show that our defenses effectively detect free-riders. 
With 1 to 5 free-riders, \defense and \defensep averagely decreases FID by 5.22\% to 11.53\% for CIFAR10 and 5.79\% to 13.22\% for CIFAR100 in comparison to an attack without defense.
In a shell, the proposed DFG(+) can effectively defend against free-riders without affecting benign clients at a negligible computation overhead.  



\end{abstract}


\section{Introduction}


Generative Adversarial Networks (GANs) are an emerging methodology to generate synthetic data ~\cite{DBLP:conf/iclr/BrockDS19,gao2018voice,zhao2021ctab,proven2021comicgan}, especially for the visual data.
 GANs are capable of generating real-world-like images 
and are increasingly adopted by the industry for data augmentation and refinement~\cite{peres2021generative}.
They owe their success to their unique architecture of two competing neural networks, called discriminator and generator.
Training GANs centrally means training a single generator and discriminator iteratively, where the former generates realistic images to fool the discriminator, and the latter then  gives feedback to the generator by comparing the generated and real images. 
As a consequence of privacy regulations imposed on data sources, e.g., GDPR~\cite{GDPR16} and HIPAA~\cite{hipaa}, 
GANs often have to employ distributed architectures such that they can learn from multiple sources without illegally sharing the raw data.

\begin{figure}[htb]
	\begin{center}
		\includegraphics[width=0.80\columnwidth]{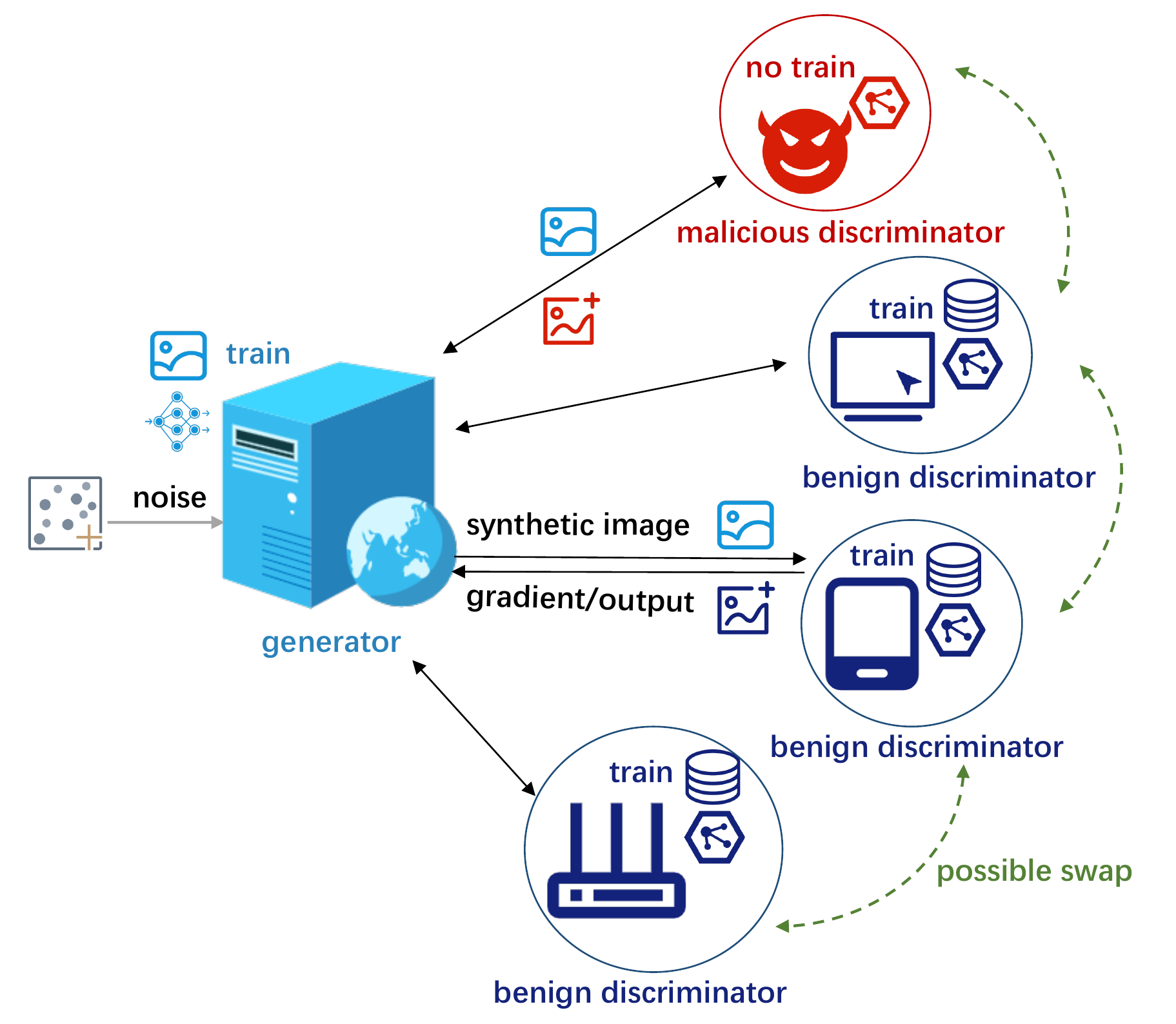}
		\vspace{-0.2em}
		\caption{Architecture of Multi-Discriminator GAN: one generator, and four discriminators, one of which being free-rider. }
		\label{fig:mdgan}
	\end{center}
	\vspace{-2em}
\end{figure}

\textbf{Multi-Discriminator GAN (\mdgan)}, 
Distributed GAN architecture have been adopted in  medical and financial domains~\cite{durugkar2016generative,asyndgan,hardy2019md,tdgan,zhao2021fed} that have stringent privacy constraints. 
Typically, as shown in Fig.~\ref{fig:mdgan}, there are one generator and multiple discriminators, one discriminator for each data source. 
To learn such an \mdgan, an iterative training procedure between generator and discriminators takes place. 
The generator synthesizes images that imitate the real data, whereas the discriminators provide feedback to the generator based on their local image set. 
Discriminators can also exchange their local networks with peers in a variant architecture of \mdgan~\cite{hardy2019md}.
Though such a distributed architecture guarantees that raw data is not shared, it increases the chance of misbehaving discriminators, such as free-riders, and increases the difficulty of defending against them.

\textbf{Free-riders} are commonly observed in distributed systems where there are multiple users participate,
e.g., peer-to-peer networks~\cite{cs5696,DBLP:journals/sigecom/FeldmanC05freep,DBLP:journals/jsac/FeldmanPCS06freep} or Federated Learning systems~\cite{DBLP:journals/tist/YangLCT19FL,DBLP:conf/aistats/McMahanMRHA17fl}. Free-riders in  Federated Learning systems~\cite{DBLP:journals/corr/freeriderlin,DBLP:conf/aistats/FraboniVL21freerider} try to gain access to the so-called global model from the server, which is aggregated from local models of all contributors without sharing local data. Here, free-riders can simply return the previous global model (possibly with perturbation added) as their contribution. 
In the context of \mdgan systems, free-riders aim to gain access to the valuable well-trained generator model without using any real data and providing any meaningful feedback to the generator. Different from the Federated Learning system where the model on the server has the identical structure of all of the clients, free-riders in \mdgan do not have any information about the global view of the generator and other discriminator networks. 
Moreover, it is no mean feat to defend against free-riders in MD-GAN as the generator only receives the distributed feedback on how well the synthetic images compared to the real ones, i.e., gradients backprogerated from the discriminator.

In this paper, we aim to answer two research questions: what is the impact of free-riders on production prototypes of \mdgan and how can benign participants defend against such free-riders? Specifically, we consider two \mdgan variants: (i) \mdsimple: discriminators only communicate with the generator~\cite{asyndgan,durugkar2016generative,tdgan}). and (ii) \mdswap: discriminators perform 
model exchanges, in a peer-to-peer way, to avoid over-fitting to their local data~\cite{hardy2019md}. We conduct the first empirical characterization study on how different the number of free-riders affects the quality of synthetic images of \mdgan when training image benchmarks.  Our results show that a small fraction of free-riders in the system can cause  degradation of the \mdgan performance, especially for \mdswap, i.e., synthetic images are highly dissimilar to the real ones, as measured by a high Fr\'echet Inception Distance (FID) score~\cite{fid}.

Secondly, we propose a novel \textbf{D}efense strategy against the \textbf{F}ree-riders on MD-\textbf{G}AN, termed DFG, where  generator can filter out the contributions of free-riders. The two key steps of DFG are (i) the generator periodically sends out a probing data set to all discriminators, and (ii) clusters  their responses in combination with the reference response of the "detector", a free-rider trained on the generator side. As the variant of \mdgan considered here allows the discriminators to periodically swap models, we further extend the defense, termed \defensep, and introduce the third defense step at the discriminators. Specifically, the discriminators cluster peers' pair-wise distance vector sent by the generator and avoid exchange updates with free-riders. Consequently, \defensep can avoid model leakage to discriminators while keeping the training benefits brought by swapping. 

We evaluate \defense and \defensep on different combinations of  percentages of free-riders and variants of \mdgan on MNIST, CIFAR10 and CIFAR-100 data sets. 
Our results show that DFG(+) can exclude the free-riders with 100\% accuracy in both variants of \mdgan even if half of the nodes in the system are free-riders. 
When we vary number of free-riders from 1 to 5, \defense and \defensep averagely decreases FID by 5.22\% (\mdsimple) and 11.53\% (\mdswap) for CIFAR10 and 5.79\% (\mdsimple) and 13.22\% (\mdswap) for CIFAR100 in comparison to an attack without defense.

 The main contributions of this paper are two fold. A first kind of characterization of free-riders on two variants of production prototypes of \mdgan is presented in Section~\ref{sec:attack}. We propose a novel and effective defense strategy \defense and its extension \defensep in section~\ref{sec:def_goals}, and evaluate them on two image benchmarks in Section~\ref{sec:experiment}. Our code is temporarily hosted on google drive\footnote{\url{https://drive.google.com/file/d/1QZxpoZpajpFf1LsLTlucl51sKPyZbKyF/view?usp=sharing}} with a detailed description for reproducing the results. 
 We plan to provide the source code via Github after publication of the paper.

\section{Free-riders on \mdgan}
\label{sec:attack}


\begin{figure*}[htb]
	\begin{center}
		\subfloat[Real MNIST images]{
			\includegraphics[width=0.16\textwidth]{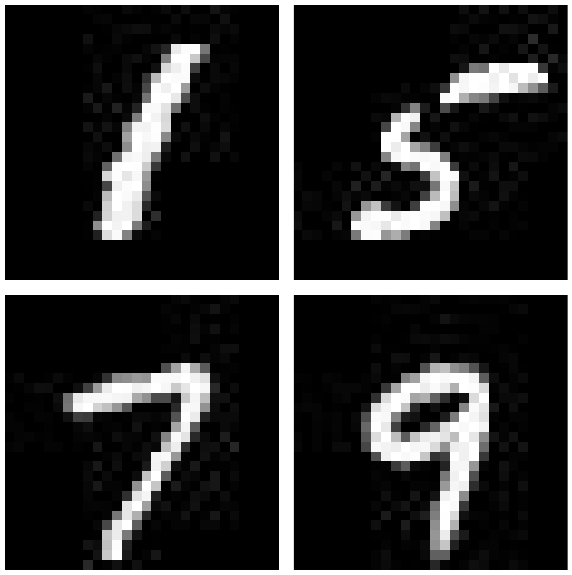}
			\label{fig:mnist_original}
		}
				\hspace{4mm}
		\subfloat[Synthetic images from \mdsimple without free-riders]{
			\includegraphics[width=0.16\textwidth]{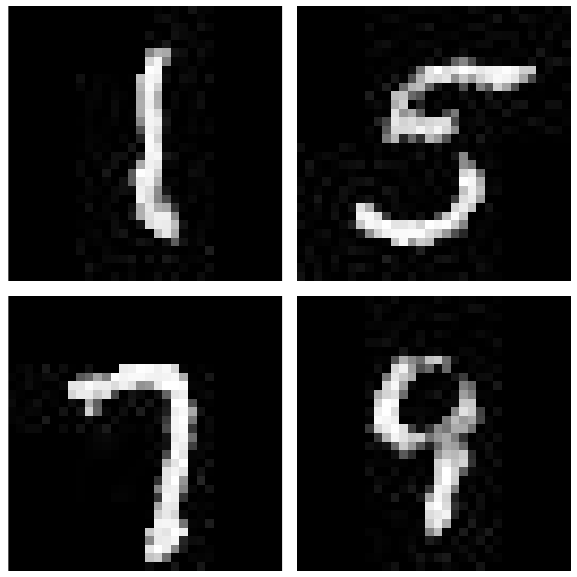}
			\label{fig:mnist_no_attacker}
		}
		\hspace{4mm}
		 \subfloat[Synthetic images from \mdsimple with 5 free-riders]{
			\includegraphics[width=0.16\textwidth]{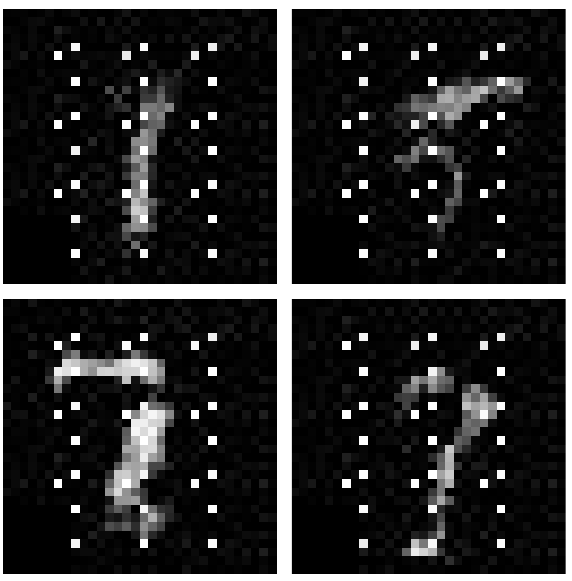}
			\label{fig:mnist_no_defense}
		}
				\hspace{4mm}
		\subfloat[FID varies with \#attacker]{
			\includegraphics[width=0.24\textwidth]{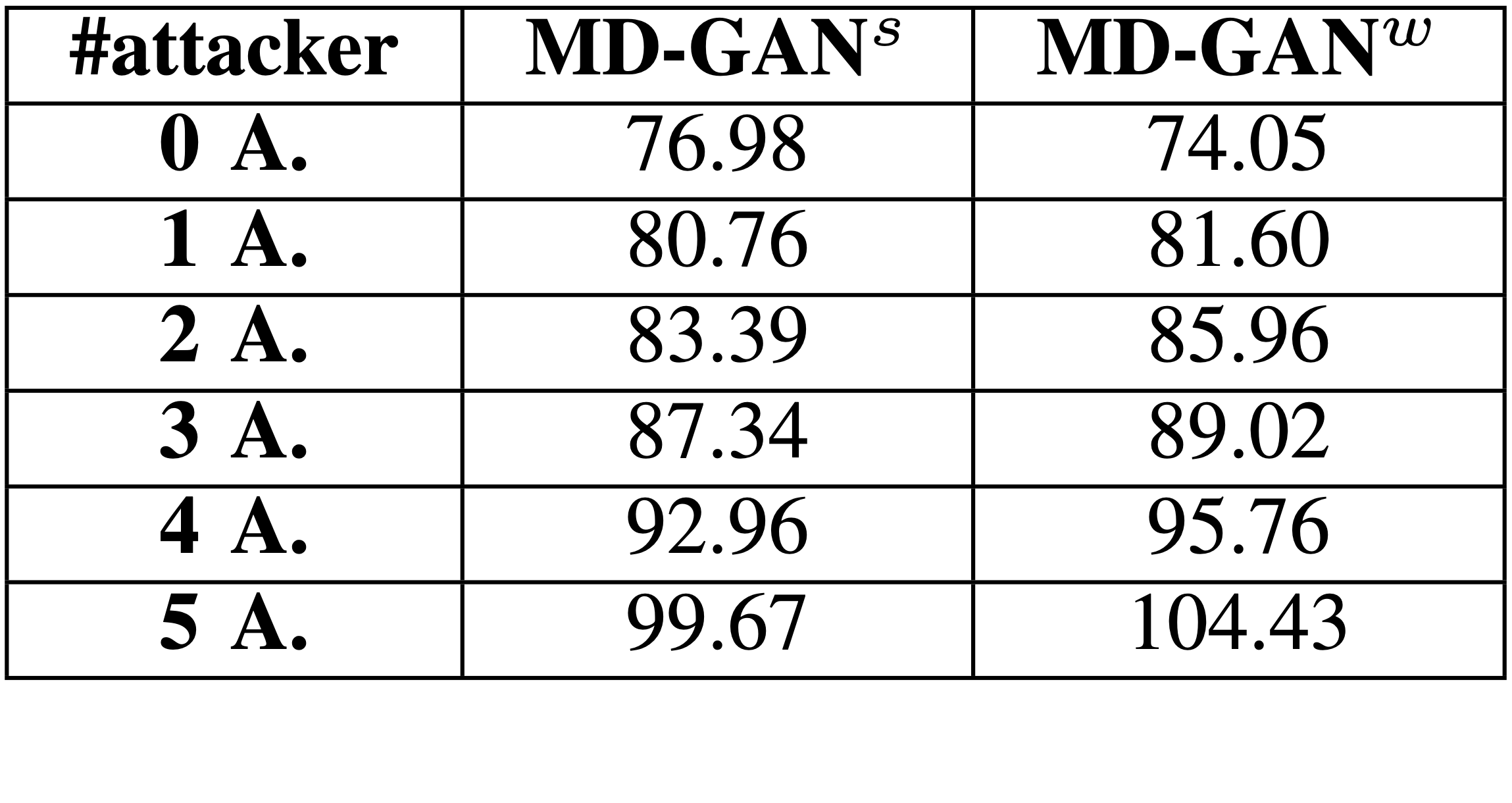}
			\label{fig:mnist_table}
		}
		\vspace{-0.em}
		\caption{Real v.s. synthetic MNIST images from generators of \mdgan encountering 0 and 5 free riders with 5 benign discriminators. The difference of FID scores across different number of free-riders.}
		\label{fig:motivation_case_mnist}
	\end{center}
	\vspace{-2em}
\end{figure*}

To quantitatively illustrate the impact of free-riders, we first introduce two architecture variants of \mdgan, which differ in their level of discriminator interaction, and present our free-rider threat model and its impact on \mdgan.

\vspace{-1em}
\subsection{System of \mdgan prototypes}

\noindent \textbf{Key components} of \mdgan are one server and $N$ clients maintaining one generator and $N$ discriminators, respectively. In general, generator and discriminators are all deep neural networks\footnote{We interchangeably use terms of networks and models} characterized by their model weights.
The generator network, $\mathcal{G}$, aims to synthesize images that are indistinguishable from real ones. On the other hand, each of the $N$ discriminator networks, $\mathcal{D}_i, i \in \{1, 2, ..., N\}$, has direct accesses to its own 
set of real images, $X_i$. 
They aim to correctly differentiate fake images generated by generator from real images. Concretely, for an image $s$, $\mathcal{D}_i(s)$ gives the probability that $s$ is real from the view of the $i$-th discriminator. 
Fig.~\ref{fig:mdgan} illustrates an example of  one generator and four discriminators.

To train an \mdgan, the generator and discriminators take turns to train and update their network weights over multiple rounds until reaching convergence. 
Following the structure of WGAN-GP \cite{wgangp}, the overall objective is composed of discriminator loss and generator loss functions, where the former captures the real/fake image classification errors
and the latter 
captures the classification mistakes on the generated images. One training round contains multiple mini batches of data. {For batch $j$ at round $t$,} 
$\mathcal{G}$ produces a synthetic dataset $S_{t,i}^j$ from an input vector of Gaussian noise $z_{t,i}^j$ for the $i^{th}$  discriminator.
    
{\bf Discriminator training}: The discriminator uses  its local real images  $X_i^j$ (i.e., real image mini batch $j$ at  $i^{th}$ discriminator) and the fake images $S_{t,i}^j$ from the generator to train itself. Specifically, the generator remains fixed during the discriminator training, we only optimize the discriminator loss and updates the weights of discriminator networks through stochastic gradient decent algorithms~\cite{robbins1951stochastic}.
    
    {\bf Generator training}: When calculating generator loss, one can imagine that generator and discriminator are connected as one neural network. 
    The { $i^{th}$ discriminator} calculates loss for fake images $S_{t,i}^j$ from the generator and back-propagates gradients. During the phase of generator training, weights of the discriminators remain fixed, only generator uses the gradients to update.
  {After $\mathcal{G}$ receives all of the back-propagated gradients of fake images $S_{t,i}^j$ from every $i^{th}$ discriminator,}
     the generator accumulates the gradients from all the discriminators and update its network weights at once.


There are two variant architectures, \mdsimple and \mdswap. In addition to iteratively training discriminators and generator networks in every round, \mdswap allows a third step of \emph{swapping discriminators' network weights}. {Every $E$ rounds, the generator randomly pairs  discriminators  to perform a peer-to-peer swapping process, i.e., exchanging their model weights. After swapping, the discriminators replace the previous model with the latest swapped model.}
Such an architecture intends to avoid over-fitting on local data for each discriminator~\cite{hardy2019md}.

\noindent\textbf{Communication model}. 
We assume the communication among the generator and discriminators is reliable and provides in-order delivery, i.e.,  messages do not get lost or re-ordered.  In practice, we use TCP to ensure both properties. 
For simplicity,  all parties are online all the time under our system and provide timely responses in each round. The generator only proceeds to its training after receiving all gradient updates from all discriminators. 
Note that it is possible to deal with offline clients~\cite{tdgan} but modeling such behavior realistically is out of the scope of this paper. 

\noindent \textbf{Implementation}
\mdgan is implemented using the Pytorch v1.8.1 RPC framework.
This choice enables the generator to control the flow of of training steps with ease. Clients just join the group, then wait to be initialized and assigned tasks. To parallelize the training across all clients, RPC provides a function \textit{rpc\_async()} that allows the generator to make non-blocking RPC calls to run functions at a client.
To implement synchronization points, RPC provides a blocking function \textit{wait()} for the return from a previous call to the function \textit{rpc\_async()}.
The return of \textit{rpc\_async()} is a \textit{future} type object. Once the \textit{wait()} is called on this object, the process is blocked until the return values are received from the client. 
We use \textit{rpc\_async()} for both training the generator and discriminators networks. 
The training of the generator network does not continue until it receives input from all discriminators and vice versa. 
For \mdswap, the swapping process also blocks the training via \textit{rpc\_async()}. 


One weakness of the current RPC framework {on} Pytorch v1.8.1 is that it does not support the transmission of 
tensor, (i.e., where all the model weights and images are loaded in memory)
directly on GPU through an RPC call. This means that each time when we collect or update the model weights, we introduce an extra delay to detach the weights from GPU to CPU or reload the weights from CPU to GPU.

\subsection{Free-rider adversarial model}

We consider free-riders on the discriminator side,  i.e., clients want to obtain the final {generator} model without contribution to the training of \mdgan. Their goal is not to degrade the image quality of the generator but the contrary.
In this sense, they are rational parties rather than malicious. They deviate from the expected learning procedure to gain utility, namely access to the generator model.   

Such free-riders are local, internal, and active adversaries.  In other words, they can only observe and participate in the communication and computation of their own training process.
Moreover, free-riders do not own any data for training \mdgan, nor do they have access to the data of others. They do not collude.  The assumption of non-collusion is sensible as additional free-riders might decrease  the quality of the final model they obtain, so parties are unlikely to reveal their free-riding to others. 
 
We specifically consider free-rider attacks with respect to \mdgan architecture variants. As discriminators participate in both the training of the discriminator and generator, free-riders take the following actions accordingly. When training the discriminator, due to lack of local data, free-riders randomly initialize the weights of their discriminator network for both variants of \mdgan. The weights follows Kaiming initialization~\cite{he2015delving}. 
{In order to provide the gradients needed in training the generator, free-riders use their randomly initialized discriminators to inference the symthetic images from generator and back-propagate.}
For \mdswap, if they manage to swap with other discriminators,  they return the gradients using the discriminator model they have received from their latest swap; otherwise, the randomly initialized model is used.


\subsection{Impact of free-riders}

 To show the impact of free-riders on \mdgan, we empirically evaluate the quality of generated images when experiencing different numbers of free-riders. We compare the generated images with the real ones using FID, the standard measure for the similarity of features between images. The lower the FID value, the better quality the generated images. The FID value is expected to decrease with the training rounds. 
 
 We performed the experiments on MNIST, CIFAR10, and CIFAR100. Due to space constraints, we only present the results for MNIST in this section. We include the result of CIFAR10 and CIFAR100 in Sec.~\ref{sec:experiment} when comparing the system with and without defense. We keep the number of benign clients constant at 5 and vary the number of free-riders between 0 and 5.  Results are averaged over 3 runs and we train for 100 rounds; with these numbers, a single run of training \mdgan takes on average around 140 minutes on our prototype. More details about the experimental setup are summarized in Sec.~\ref{ssec:setup}.
 
 Fig.~\ref{fig:motivation_case_mnist} displays the impact of free-riding  through both illustration and concrete values. Fig.~\ref{fig:mnist_no_defense} and Fig~\ref{fig:mnist_no_attacker} highlight the quality difference between the synthetic images when encountering 0 and 5 free-riders, respectively, for \mdsimple. Free-riders clearly degrade the capacity of \mdsimple in generating realistic images. 
 This intuition is quantified by the FID in Fig.~\ref{fig:mnist_table}. Furthermore, 
 when there are no free-riders, \mdswap converges better than \mdsimple. That is because swapping prevents discriminators from  over-fitting on their local data. With the increasing number of free-riders in the system, the performance of \mdsimple and \mdswap degrades. \mdswap deteriorates more than \mdsimple for same number of free-riders, because swapping with free-riders destroys all the previous training effort of a benign client. The result illustrates that free-riders, though they do not intend to negatively affect the final result, have a concerning negative impact on performance, emphasizing the need for defenses.  
\section{Defending \mdgan against Free-riders}
\label{sec:def_goals}

After introducing models and goals,  we propose \dft (i.e., \defense and \defensep), a defense strategy against free-riders in \mdgan. The key elements of \dft are a probing generated image set $\hat{S}$ and a \detector $\mathcal{D}_{N+1}$. 
$\mathcal{D}_{N+1}$ is a free-rider run by the generator to obtain a reference model for free-rider responses. 
As \mdswap allows discriminators to swap local model weights, we further enhance the defense for the discriminators, termed \defenseplus. 
\defenseplus provides discriminators with information about the similarity across their responses on the probing set $\hat{S}$ 
 and enables them to strategically accept or reject swaps.   
Fig.~\ref{fig:defense_scheme} illustrates the steps and the differences in the defenses. 


\noindent The \textbf{objectives of the \defense} are three-fold:
\begin{itemize}
\item[\textbf{1}] Accurately detecting free-riders in each round and excluding their gradients from accumulation. 
\item[\textbf{2}] Improving the FID for the case when free-riders are present but not considerably decreasing the FID in the absence of free-riders.
\item[\textbf{3}] Entailing low additional overhead. 
\end{itemize}

Note that the first goal also implies that benign clients should not be classified as free-riders.  The second part of the second goal is important as a defense that decreases the performance,  e.g., by excluding benign clients,  in the absence of an attack is unlikely to be adopted, especially if the impact of a low number of free-riders is less than the decrease in image quality caused by the defense. 
The last goal is particularly important because the generator and discriminators might be unwilling to deploy a defense that considerably increases delays, computation, or communication overhead. 

We analyze the overhead of the defense theoretically at the end of the section and provide experimental results to confirm that we achieve the first two objectives in Section~\ref{sec:experiment}.

\vspace{-0.5em}
\subsection{Protocol of \defense}
\vspace{-0.5em}

\textbf{Step 1:} In our defense, $\mathcal{G}$ periodically, i.e., every $L$ rounds,  generates a probing set $\hat{S}$ to the clients. The set can act as a replacement for $S_{t,i}^j$ (i.e., synthetic image at round $t$ and batch $j$ of the $i^{th}$ discriminator).
In contrast to the standard algorithm,  \defense sends the same set $\hat{S}$ to all clients.  The clients evaluate their discriminators on the set $\hat{S}$ and return the results in the form of a vector.  Concretely,  for each image $s_k$, with $1\leq k \leq |\hat{S}|$, discriminator $\mathcal{D}_{i}$ computes 
$\mathcal{D}_{i}(s_k)$ 
and the returned vector is:
\begin{equation*} 
Pr_i(\hat{S}) = \left(\mathcal{D}_{i}(s_1),  \mathcal{D}_{i}(s_2), \ldots , \mathcal{D}_{i}(s_{|\hat{S}|})\right).
\end{equation*} 

\textbf{Step 2:} Additionally,  to model how free-riders behave,  $\mathcal{G}$ makes use of a \detector. The usage of the \detector is sharing the same thought as the usage of oracle as a reference in active learning for defending attacks ~\cite{Zhao:DSN19,Zhao:TDSC21, younesian2021qactor}. Concretely,  the generator randomly initializes a discriminator $\mathcal{D}_{N+1}$ as a reference model of a free-rider and then computes $Pr_{N+1}(\hat{S})$. 

\textbf{Step 3:} After the generator collects all the vectors $Pr_i$,  $1\leq i \leq N+1$, it applies 2-means clustering based on the distances of the squared L2 norm
on all vectors $Pr_i$. 
 Intuitively, the $Pr_i$ of a benign client is expected to have a low distance to the $Pr_i$ of other benign clients, whereas they have a high distance to the $Pr_i$ of the free-riders, including the \detector. Consequently, we classify all clients in the cluster that contains the \detector as free-riders.

\vspace{-0.5em}
\subsection{Protocol of \defensep}
\vspace{-0.5em}
\mdswap provides both additional challenges and opportunities.  A key challenge is that a discriminator is not trained by one single client and hence it is hard to determine whether one party has trained properly. 
Free-riders can obtain a properly trained discriminator by swapping, and then compute the gradients of images from the generator. This further exacerbate the difficulty of differentiating the gradients obtained from free-riders and benign discriminators. 
To introduce a discriminator-side defense, we take advantage of one information: the benign discriminators know that they are not  free-riders. So once a benign client is asked to swap with another that is suspected to be a free-rider, it can refuse.

\defenseplus utilizes this extra information to guide the swapping process by adding three more steps to \defense. 

\textbf{Step 4:} After the generator has all the vectors $Pr_i$,  $1\leq i \leq N+1$,  they compute a $(N+1)\times(N+1)$  matrix $V$ of pair-wise L2 distances between the $Pr$ of the discriminators, including the detector, i.e.,  the element $V_{ij}$ is $||Pr_i - Pr_j||_2$. 
The generator shares the computed distances $V_{i1}, \ldots V_{i(N+1)}$ with the $i^{th}$ client.

\textbf{Step 5:} A benign client $i$ then performs 2-means clustering on these distances, excluding $V_{ii}$. 
The cluster with lower mean distances is taken to be the cluster of benign clients. The underlying assumption here is that the differences between properly trained discriminators is less than between randomly initialized ones. 

\textbf{Step 6:} A benign client only swaps with parties that are in the same cluster as it according to its local clustering. 

While our protocol cannot dictate the behavior of free-riders (as they may choose not to follow the protocol), in our evaluation, free-riders always accept swap requests as they hope to obtain better models. 


\subsection{Overhead}
\vspace{-0.5em}
We discuss the computation and communication complexity of \defense and \defensep in relation to the complexity of MD-GANs without any defense. 
We start by analyzing the additional computation overhead, which is the main cost factor as by our experiments. 

\noindent\textbf{Computation \defense:} In \defense, note that the clients need to compute $D_i(s_k)$ for $s_k \in \hat{S}$. However, when training without the defense, they also need to compute $D_i(s'_k)$ for $s'_k \in S_{t,i}^j$ with $|S_{t,i}^j|=|\hat{S}|$
to evaluate the quality of their current model. Hence, there is no additional overhead on the client-side with regard to computation. The generator has to compute the pair-wise distances between $N+1$ vectors of length $|\hat{S}|$  and perform a 2-means clustering of $N+1$ values. The distance computation has complexity $\mathcal{O}\left(|\hat{S}|N^2\right)$ and the k-means clustering has complexity $\mathcal{O}\left(N^2\right)$. As the number of images is expected to be much larger than the number of clients, we have a complexity of $\mathcal{O}\left(|\hat{S}|N^2\right)$. In comparison, the cost of a training round without \defense is $\mathcal{O}\left(|\hat{S}|MQ\right)$ 
where $M$ is the image size and $Q$ a factor related to the structure and size of the neural network~\cite{e20040305}. 
According to the real world application of \mdgan~\cite{asyndgan}, $M$ should be much larger than $N^2$, meaning that the computation complexity of the normal \mdgan execution by far exceeds the complexity of \defense. 

\noindent\textbf{Computation \defensep:} For \defensep, the clustering is also performed on the client-side, incurring an overhead of $\mathcal{O}\left(N^2\right)$ for each client. 
As for \defense, the training complexity exceeds the complexity of \defensep by far. 

\noindent\textbf{Communication \defense:} The only additional communication overhead is on the client-side, namely sending the values $D_i(s_k)$ to the generator. Hence the communication complexity is $\mathcal{O}(N|\hat{S}|)$. During a normal \mdgan round, the generator sends $N|\hat{S}|$ images of size $M$, so the complexity is $\mathcal{O}(MN|\hat{S}|)$, much higher than for \defense. 

\noindent\textbf{Communication \defensep:}
For \defensep, the generator sends $N$ vectors of length $N+1$ that allow the clients to perform clustering, so the communication complexity of \defensep is $\mathcal{O}(N|\hat{S}|+N^2)$. In practice, we expect a low number of clients in comparison to images, so $\mathcal{O}(N|\hat{S}|)$ is the dominating factor, which, as explained for \defense, is much lower than the complexity of an \mdgan execution without defense. 

Overall, the complexities of \defense and \defensep are much lower than those of the standard training, so that they do not add overheads that might discourage use of the defenses. 




\begin{figure}[t]
	\begin{center}
	{
			\includegraphics[width=0.85\columnwidth]{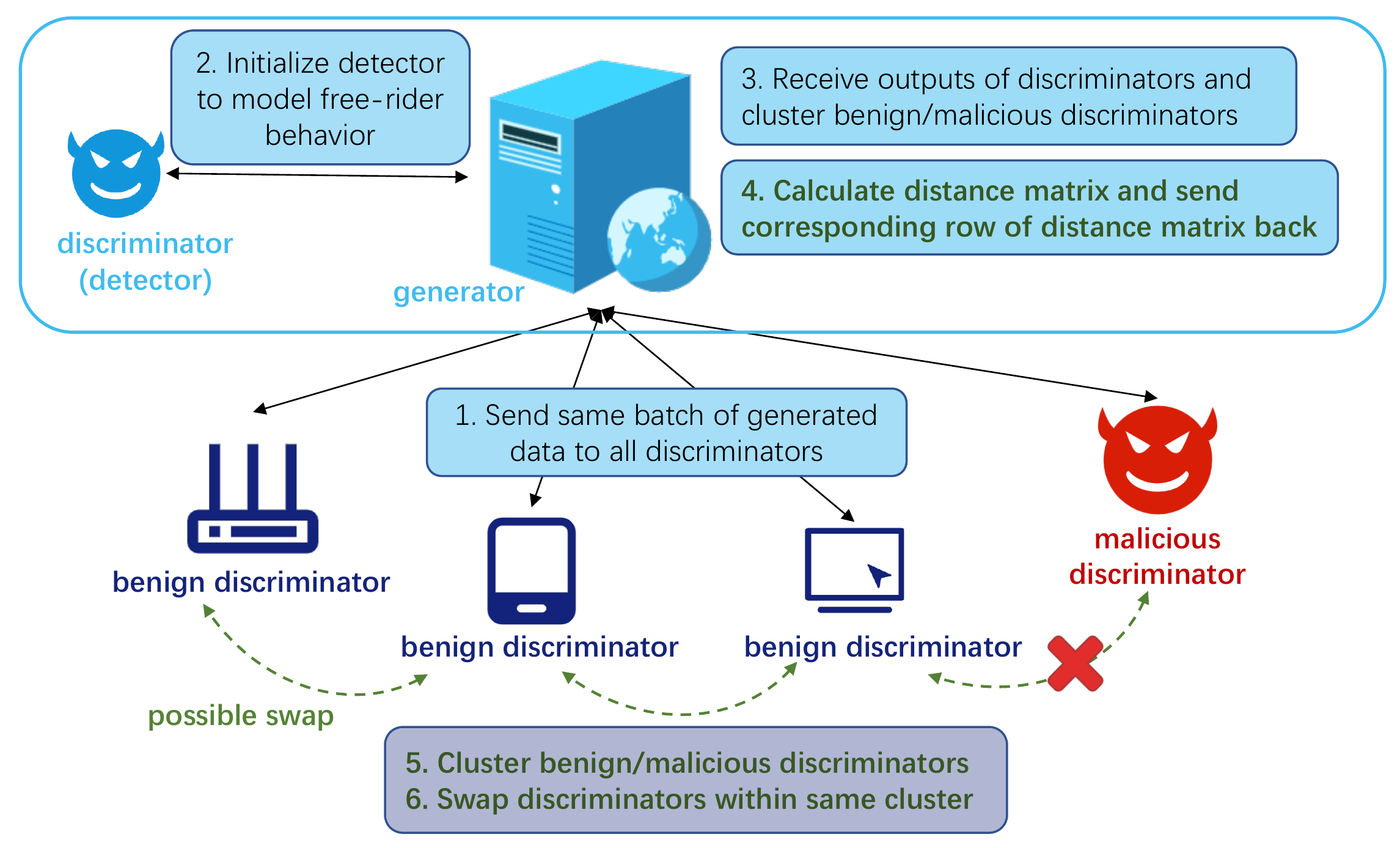}
			\label{fig:both_side_defense}
		}

		\caption{Key steps of DFG(+): step 1-3 are common to \mdsimple and \mdswap, whereas step 4-6 are specific to \mdswap only.} 
		\label{fig:defense_scheme}
 	\end{center}
 	\vspace{-1.2em}
\end{figure}
\vspace{-0.5em}
\section{Experimental Evaluation}
\label{sec:experiment}
Our defense \defense and \defensep are evaluated on three widely used image datasets, and compared with the scenarios without defense. To evaluate the efficiency of \defense, we resort to the precision and recall of the identified free-riders. And for \defensep, we also report how many malicious swap requests our defense prevents. We furthermore conduct an experiment without the \detector to show-case its importance to correctly identify free-riders.

\subsection{Evaluation Metrics}
\label{ssec:evaluation_metrics}
To evaluate all the experiments, we test the final performance of $\mathcal{G}$. Since our use case is image classification, we choose to use Fr\'echet inception distance (FID)~\cite{fid} defined as follows:
\begin{equation*}
    \mbox{FID} = ||\mu_1-\mu_2||^2 + \operatorname{tr}(\Sigma_1+\Sigma_2-2(\Sigma_1\Sigma_2)^{1/2})
\end{equation*}
where  $\mu_1$ and $\mu_2$ denote the feature-wise mean of the real and generated images, $\Sigma_1$ and $\Sigma_2$ refer to the covariance matrix for the real and generated feature vectors.  $||\mu_1-\mu_2||^2$ refers to the sum squared difference between the two mean vectors, and $tr$ is the trace linear algebra operation. 
Intuitively, the lower the FID, the closer the generated and real images.

To show the efficiency of \defense and \defensep, we use two different metrics. For \defense, 
the \textbf{precision} and \textbf{recall} of the identified "free-riders"  
are reported. {The precision quantifies the fraction of predicted "free-riders" are actual real free-riders. The recall is to measure the fraction of true free-riders identified by our defense.} Free-rider and benign client are labelled as Positive and Negative for the calculation ~\cite{DBLP:journals/corr/pr}.
Note that recall is not defined in the absence of free-riders. 

For \defensep, our focus lies in preventing discriminator swapping between benign and malicious clients. If benign clients swap with benign clients or malicious clients swap with malicious clients, we regard it as correct action. But if the \defensep prevents a swapping request between two benign clients, we see this as a \textbf{wrong prevention}. And if \defensep does not stop a  swapping between a benign and a malicious client, we call this a \textbf{wrong permission}. We count all the actions above and report their relative frequency.

\subsection{Experimental setup}
\label{ssec:setup}
\textbf{Datasets.} We test our algorithms on three commonly used image datasets: MNIST\cite{lecun:1998:mnist}, CIFAR10\cite{krizhevsky:2009:cifar} and CIFAR100\cite{krizhevsky:2009:cifar}. MNIST contains 60 000 grayscale (10 classes) while CIFAR10 and CIFAR100 have 50 000 colorful (10/100 classes) training images. Each benign client possesses 5 000 images, which are evenly distributed over all of the classes.

\textbf{Baselines.} To show the efficiency of \defense and \defensep, we run the algorithms \mdsimple and \mdswap with (i.e., \textbf{DFG$\_$Simple} and \textbf{DFG+$\_$Swap}) and without (i.e., \textbf{\mdsimplenodefense} and \textbf{\mdswapnodefensep}) defense. We fix the number of \textbf{benign clients to 5} for all experiments and vary the number of malicious clients (i.e., free-riders) from 0 to 5.  The total number of training rounds is 100.
$\mathcal{G}$ generates 10 000 images every 5 rounds and they are used to evaluate $\mathcal{G}$'s performance in terms of FID. 
 Every 10 rounds, we execute \defense and \defensep: the  generator sends the same probing set $\hat{S}$ of 500 images to all clients and the detector, though $\hat{S}$ varies over rounds.

For all experiments, we choose to use Wasserstein GAN with Gradient Penalty (WGAN-GP)~\cite{wgangp} structure as generator and discriminator models. {The network of each discriminator consists of three repeated blocks. Each block  concatenates 2D Convolution, Instance Normalization and Leaky Relu layers. $\mathcal{G}$ is also composed of three concatenating blocks. Each block contains 2D Transposed Convolution, Batch Normalization and Relu layers.}
The batch size $B$ is set to 500. Since each client owns 5 000 images, there are 10 mini batches per training round. Due to the characteristics of WGAN-GP, the generator is trained once for every 5 times the discriminators are trained. Therefore, for each round, the discriminator is trained 10 times by all mini batches, but the generator is only trained twice. For \defense and \defensep, when they evaluate the quality of discriminator every 10 rounds, they only do that during the first training batch out of two within the round. We repeat each experiment 3 times and report the average.


\textbf{Testbed.} Experiments are run on two machines, both running Ubuntu 20.04. Each machine is equipped with 32 GB memory, GeForce RTX 2080 Ti GPU and 10-core Intel i9 CPU. Each CPU core has two threads, hence each machine contains 20 logical CPU cores in total. One machine hosts the generator, the other hosts all the discriminators. The machines are interconnected via 1G Ethernet links.

\vspace{-0.5em}
\subsection{Results}
\vspace{-0.5em}
\subsubsection{\defense and \defensep for \mdsimple and \mdswap}
Tab.~\ref{table:final_fid} shows the final FID of \mdsimple and \mdswap with and without defenses (i.e., \defense and \defensep). \defense and \defensep improve the final performance of the generator. 
The \defensep improvement for \mdswap is higher than the improvement made by \defense for \mdsimple. 
That is because, as detailed in Section~\ref{sec:attack}, the impact of the attack is greater on \mdswap, so there is more room for improvement. 
With 1 to 5 free-riders, \defense and \defensep averagely decrease FID by 5.22\% (\mdsimple) and 11.53\% (\mdswap) for CIFAR10 in comparison to an attack without defense. For CIFAR100, the decrease is even slightly higher, namely 5.79\% (\mdsimple) and 13.22\% (\mdswap). 

For all the experiments on CIFAR10 and CIFAR100, the precision and recall of the recognition of free-riders by \defense are \textbf{100\%}. That means for the experiments with free-riders, \defense correctly identifies all the attackers. And for the experiments without free-riders, \defense never wrongly excludes any benign client. For \defensep, the ratio of correct preventions and permissions is also \textbf{100\%} for all the experiments. It signifies that all the swaps are between two benign clients or two free-riders, but never between a free-rider and a benign client.
Thus, Objective 1 from Section~\ref{sec:def_goals}, i.e., correctly identifying free-riders, is achieved. 
There are no bold results for 0 attacker in Tab.~\ref{table:final_fid} because there is no statistically significant difference between having a defense and not having a defense. 
This result shows that our defense does not negatively impact FID, thus fulfilling Objective 2 from Section~\ref{sec:def_goals}.  

The improvement from the defenses is more pronounced when the number of attackers is high. 
With an increasing number of free-riders, the harm accumulates in the system. But with our defenses, all the free-riders are correctly identified and excluded. The only damage for the system comes from the first 10 rounds before our defenses are applied.

\begin{table*}[h]
\centering
\caption{Final FID for \mdsimple and \mdswap (\textbf{A.} is short for free-rider attacker).}
\label{}
\resizebox{0.9\textwidth}{!}{
\begin{tabular}{|c||c|c|c|c|c|c||c|c|c|c|c|c|}
\hline
\multirow{2}{*}{\textbf{Method}} & \multicolumn{6}{c||}{\bf CIFAR100} &
\multicolumn{6}{c|}{\bf CIFAR10} \\
\cline{2-13}
 & \textbf{0 A.}& \textbf{1 A.} & \textbf{2 A.}  & \textbf{3 A.}& \textbf{4 A.}&\textbf{5 A.}  & \textbf{0 A.}& \textbf{1 A.} & \textbf{2 A.}  & \textbf{3 A.}& \textbf{4 A.}&\textbf{5 A.}  \\
\hline
No\_Def\_Simple & 80.45 &85.54&86.14&87.62&92.34&96.07
&78.59&81.71&84.86&86.06&92.01&95.67\\
\hline
DFG\_Simple &80.98 &\textbf{81.36}&\textbf{82.98}&\textbf{85.01}&\textbf{86.43}&\textbf{88.21}&78.15&\textbf{80.32}&\textbf{81.25}&\textbf{82.34}&\textbf{84,15}&\textbf{85.96}\\
\hline
No\_Def\_Swap &79.40 &89.83&92.51&95.52&99.28&101.37& 78.36 &	90.36&	92.55&	95.70&	98.09&	101.98\\
\hline
DFG+\_Swap &79.35&\textbf{79.69}&\textbf{82.94}&\textbf{84.62}&\textbf{85.69}&\textbf{90.31}& 78.83&\textbf{80.31}&\textbf{81.45}&\textbf{81.89}&\textbf{85.46}&\textbf{86.01}\\
\hline
\end{tabular}
}
\label{table:final_fid}
\end{table*}

\subsubsection{Ablation study}
To show the efficacy of the detector, we propose an ablation experiment without the detector. 
Therefore, we adjust \defense to \defenseadjust: after step 3 in Fig.~\ref{fig:defense_scheme}, \defenseadjust calculates the pairwise distances of the $N$ clients without the detector and then calculates the sum of distances for each discriminator. 
To determine free-riders, \defenseadjust  uses 2-means to cluster the sum of distances. The group with the higher mean is treated as the cluster of free-riders. The motivation behind this design is based on the conjecture that outputs from benign clients should be similar and far from free-riders. For adapted \defenseadjustp, the only difference to \defensep is that there is no distance to the detector in the vector provided to the clients. 
Due to the page limit, we only present the experiments with 0 and 5 attackers. The setup is the same as in Sec.~\ref{ssec:setup}. 

Tab.~\ref{table:ablation_simple} and Tab.~\ref{table:ablation_swap} compare \defenseadjust and \defenseadjustp with \defense and \defensep for both \mdsimple and \mdswap. 
The values in Tab.~\ref{table:ablation_simple} and Tab.~\ref{table:ablation_swap} are the absolute difference of the various metrics between \defense/\defenseadjust and \defensep/\defenseadjustp, respectively. For instance, the value $-12.77\%$ in the third row and second column of Tab.~\ref{table:ablation_simple} indicates that the FID of \defense is 12.77\% lower --- and hence better --- than the one of \defenseadjust.  
 \defenseadjust and \defenseadjustp wrongly exclude benign clients if there is no attacker. 
 This degradation in the absence of attacks happens
because the clustering always divides the clients in two clusters, with one of them being treated as free-riders. In the absence of actual free-riders, benign clients are excluded.
\defenseadjust also excludes benign clients for the case of 5 attackers, indeed it is more likely to exclude benign clients than free-riders. \defenseadjustp fares slightly better for the case of 5 attackers but still misclassifies some benign clients as free-riders and vice versa. 
The misclassification happens when the distances between different free-riders are similar to those between benign clients, so that the generator assumes the wrong cluster to consist of free-riders. 
The experiment illustrates the need for a detector as a reference point. 

\begin{table}[htb!]
\centering
\caption{Positive impact of detector for \defense on \mdsimple}
\label{}
\resizebox{0.9\columnwidth}{!}{
\begin{tabular}{|c||c|c||c|c|}
\hline
\multirow{2}{*}{\textbf{Metrics}} & \multicolumn{2}{c||}{\bf CIFAR100} &
\multicolumn{2}{c|}{\bf CIFAR10} \\
\cline{2-5}
 & \textbf{0 A.} &\textbf{5 A.}  & \textbf{0 A.}&\textbf{5 A.}  \\
\hline
FID & -12.77\% & -10.44\%  &-13.81\%&-15.41\%\\
\hline
Precision &+100\% &+51.18\%&+100\%&+79.20\%\\
\hline
Recall &- &+73.34\%&-&+88.89\%\\
\hline
\end{tabular}
}
\label{table:ablation_simple}
\vspace{-0.5em}
\end{table}

\begin{table}[htb!]
\centering
\caption{Positive impact of detector for \defensep on \mdswap}
\label{}
\resizebox{0.9\columnwidth}{!}{
\begin{tabular}{|c||c|c||c|c|}
\hline
\multirow{2}{*}{\textbf{Metrics}} & \multicolumn{2}{c||}{\bf CIFAR100} &
\multicolumn{2}{c|}{\bf CIFAR10} \\
\cline{2-5}
 & \textbf{0 A.} &\textbf{5 A.}  & \textbf{0 A.}&\textbf{5 A.}  \\
\hline
FID & - 7.07\%  & - 8.58\% & - 20.58\%& -7.79\%\\
\hline
Precision &+100\% &+44.83\%&+100\%&+75.54\%\\
\hline
Recall &- &+64.45\%&-&+47.62\%\\
\hline
Correct Prev. \& Perm.&0\%&0\%& 0\%& 0\%\\
\hline
Wrong Prevention&-48.15\%&0\%&-88.89\%&0\%\\
\hline
Wrong Permission&0\%&0\%&0\%&0\%\\
\hline
\end{tabular}
}
\label{table:ablation_swap}
\end{table}
\section{Related work}
\label{sec:related}
In this section, we summarize the related studies on multi-discriminator GAN frameworks~\cite{durugkar2016generative,hardy2019md,asyndgan,tdgan} and free-riders attacks in distributed learning systems~\cite{DBLP:conf/aistats/FraboniVL21freerider,DBLP:journals/corr/freeriderlin}. 

\textbf{\mdgan: }Overcoming the data privacy issues of centralized GANs~\cite{DBLP:conf/dsn/LiuKK19gan,DBLP:conf/iccv/LiangHZGX17gan,DBLP:conf/iclr/LiAXJS18dgan}, distributed GANs~\cite{durugkar2016generative,hardy2019md,asyndgan,tdgan,DBLP:conf/middleware/GuerraouiGKM20fegan,DBLP:journals/corr/fedgan}  enable multiple data owners to collaboratively train GAN systems. Existing distributed GAN frameworks can be summarized as Federated Learning GANs (FLGANs)~\cite{DBLP:conf/middleware/GuerraouiGKM20fegan,DBLP:journals/corr/fedgan} and MD-GANs~\cite{durugkar2016generative,hardy2019md,asyndgan,tdgan}. In FLGANs, a client trains both a generator and a discriminator network and a server aggregates both networks from all clients. Specifically, FedGAN~\cite{DBLP:journals/corr/fedgan} addresses the issue of heterogeneous data sources subject to communication and privacy constraints and provides a convergence guarantee. FeGAN~\cite{DBLP:conf/middleware/GuerraouiGKM20fegan} then focuses on the system view to overcome mode collapse and learning divergence problems arising from a distributed setup. Its main contributions are Kullback-Leibler (KL) weighting and a balanced sampling scheme. Training an entire GAN is computation intensive, especially for the generator. Thus, the implicit prerequisite for FLGANs is that clients have sufficient computational capacity.   
In contrast, \mdgan architectures offload the intensive training of the generator to the server and keep the lighter training of the discriminator on the client side. 
In this manner, MD-GANs are also able to involve a massive number of  edge nodes~\cite{DBLP:conf/infocom/WangXLHQZ21edge,DBLP:conf/dsn/Correia0R20edge}. 
The various architectures of \mdgan differ with regard to model exchange between discriminators. 
AsynDGAN\cite{asyndgan} and GMAN~\cite{durugkar2016generative} are elementary \mdgan architectures where discriminators 
only directly communicate with the generator. 
In order to improve the drawbacks of \mdgan when discriminators only own small datasets, Hardy et al.~\cite{hardy2019md} propose that discriminators are swapped between clients, opening an opportunity for free-riders to act stealthily.

\textbf{Free-riders: }The concept of free-riders
first emerged in economics~\cite{baumol2004welfare} 
but has been essential in various of distributed systems. In peer-to-peer file-sharing systems, free-riders join to download files without uploading any files~\cite{DBLP:conf/hotnets/LocherMSW06freep,DBLP:journals/sigecom/FeldmanC05freep}. 
In Federated Learning systems~\cite{DBLP:journals/tist/YangLCT19FL,DBLP:conf/aistats/McMahanMRHA17fl},  Lin et. al.~\cite{DBLP:journals/corr/freeriderlin} first propose stealthy free-rider attacks for image classification: instead of sending a random model, free-riders send the global model of the previous round back with small perturbation noises added or provide a fake gradient using the previous difference of weights. Defenses are designed accordingly based on the DAGMM~\cite{DBLP:conf/iclr/ZongSMCLCC18DAGMM} network, which is a recent anomaly detection method so as to catch the differences on deep feature by gradients for free-riders. Fraboni et. al.~\cite{DBLP:conf/aistats/FraboniVL21freerider} further explore the attack of adding perturbation noises of \cite{DBLP:journals/corr/freeriderlin} to provide convergence guarantee of the global model in the presence of a single free-rider. This work establishes the theoretical support for global convergence using a similar attacking strategy. However, as both studies are concerned with Federated Learning systems, where the clients and the server are curating models of the same structure, they are not directly applicable to \mdgan systems where the server and client train different types of models. 
Additionally, none of them has provided a systematic study on the influence of (multiple) free-riders.
To the best of our knowledge this paper is the first to study free-riders in MD-GANs.

\section{Conclusion}
In this paper, we first showed that MD-GANs suffer noticeably from free-rider attacks, especially for \mdswap where the discriminators exchange local models. With 1 to 5 free-riders, our defense, \dft, averagely increases the quality of the synthetic images by 5.22\% (\mdsimple) and 11.53\% (\mdswap) for CIFAR10 and 5.79\% (\mdsimple) and 13.22\% (\mdswap) for CIFAR100.  

In our experiments, the defenses are always successful in classifying free-riders and benign nodes when a \detector is present.  However,  while all free-riders initialize different random models,  they all use the same methodology for initialization as the \detector. If that is not the case, e.g., when free-riders are aware of the defense based on the detector,  free-riders are still expected to perform very differently than benign nodes and hence we do not expect them to be misclassified.  It is future work to verify this expectation and ensure that defenses can mitigate free-riders if these free-riders are aware of the defense. 
\bibliographystyle{abbrv}
\bibliography{reference}


\end{document}